\documentclass[10pt,twocolumn,letterpaper]{article}

\usepackage[pagenumbers]{iccv} 

\definecolor{iccvblue}{rgb}{0.21,0.49,0.74}
\usepackage[pagebackref,breaklinks,colorlinks,allcolors=iccvblue]{hyperref}

\usepackage{graphicx}
\usepackage{amsmath}
\usepackage{amssymb}
\usepackage{booktabs}
\usepackage{multirow}
\usepackage{makecell}
\usepackage{subcaption}
\usepackage{color}
\usepackage{tcolorbox}
\usepackage{pifont}
\usepackage{amsthm,lipsum}
\usepackage{siunitx}
\usepackage{enumitem}
\usepackage{algorithm}
\usepackage{algorithmic}
\usepackage{newfloat}
\usepackage{listings}
\usepackage{arydshln}
\usepackage{mathbbol}
\usepackage[accsupp]{axessibility}  
\DeclareMathOperator*{\argmax}{arg\,max}
\DeclareCaptionStyle{ruled}{labelfont=normalfont,labelsep=colon,strut=off} 
\floatstyle{ruled}
\newfloat{listing}{tb}{lst}{}
\floatname{listing}{Listing}

\def\astar{A$^\ast$}
\def\tstar{Theta$^\ast$}
\newcommand{\mb}[1]{\mathbf{#1}}
\newcommand{\mc}[1]{\mathcal{#1}}
\newcommand{\acc}[2]{\prlentwo{#1}$\pm$\scriptsize{\prlenone{#2}}}

\ExplSyntaxOn
\NewDocumentCommand{\prlenone}{m}{%
  \num[
    detect-all,
    round-mode=places,
    round-precision=1,
    ]{\fp_eval:n {{\ifdim #1pt>0pt (\ifdim #1pt>0.1pt #1pt\else 0.1pt\fi) \else #1pt\fi}}}
 }
 
\NewDocumentCommand{\prlentwo}{m}{%
  \num[
    detect-all,
    round-mode=places,
    round-precision=2,
  ]{\fp_eval:n { \dim_to_fp:n {#1pt}}}
 }
\ExplSyntaxOff

\newcommand{\iccvcolor}[1]{{\color{iccvblue}{#1}}}
\newcommand{\bestcolor}[1]{\textbf{#1}}
\newcommand{\secbestcolor}[1]{\underline{#1}}
\theoremstyle{definition}
\newtheorem{remark}{Remark}
\theoremstyle{definition}
\newtheorem{definition}{Definition}

\def\confName{ICCV}
\def\confYear{2025}

\title{DA\astar: Deep Angular A Star for Image-based Path Planning}

\author{
Zhiwei Xu\stepcounter{footnote}\thanks{Parts of this work were done at The Australian National University.}\\
School of Computing and Information Systems\\
The University of Melbourne, Australia\\
{\tt\small danny.xu@unimelb.edu.au}
\vspace{-5mm}
}

\begin{document}

\maketitle

\begin{abstract}
Path smoothness is often overlooked in path imitation learning from expert demonstrations. In this paper, we introduce a novel learning method, termed deep angular \astar~(DA\astar), by incorporating the proposed path angular freedom (PAF) into \astar~to improve path similarity through adaptive path smoothness. The PAF aims to explore the effect of move angles on path node expansion by finding the trade-off between their minimum and maximum values, allowing for high adaptiveness for imitation learning. DA\astar~improves path optimality by closely aligning with the reference path through joint optimization of path shortening and smoothing, which correspond to heuristic distance and PAF, respectively. Throughout comprehensive evaluations on 7 datasets, including 4 maze datasets, 2 video-game datasets, and a real-world drone-view dataset containing 2 scenarios, we demonstrate remarkable improvements of our DA\astar~over neural \astar~in path similarity between the predicted and reference paths with a shorter path length when the shortest path is plausible, improving by \textbf{9.0\% SPR}, \textbf{6.9\% ASIM}, and \textbf{3.9\% PSIM}. Furthermore, when jointly learning pathfinding with both path loss and path probability map loss, DA\astar~significantly outperforms the state-of-the-art TransPath by \textbf{6.3\% SPR}, \textbf{6.0\% PSIM}, and \textbf{3.7\% ASIM}. We also discuss the minor trade-off between path optimality and search efficiency where applicable.
\end{abstract}

\vspace{-5mm}

\section{Introduction}

Path planning~\cite{ppp_1,ppp_2} is a fundamental technique in various fields, including robotics and computer vision for video game~\cite{video_games_1,video_games_2,video_games_3},
visual navigation~\cite{random_tree,EBSAStar},
geodesic distance measurement~\cite{3d_app},
autonomous driving~\cite{water_traffic,auto_land_vehicle},
unmanned aerial vehicle routing~\cite{UAV_multi_obj,RJAStar,gradsmqga},
and smart agriculture~\cite{farm_1,farm_2}.
It finds the optimal route between the source and destination in 2D and 3D environments~\cite{3d_1,3d_2} with spatial obstacles or impenetrable areas.

Traditional pathfinding algorithms, such as depth-first search~\cite{depth_first},
breadth-first search~\cite{breadth_first}, and Dijkstra's algorithm~\cite{dijkstra},
use greedy search strategies across all active nodes, often resulting in high computational complexity.
To mitigate this, heuristic path planning methods, including \astar~\cite{astar} and Theta$^\ast$~\cite{theta_star}, leverage heuristic guidance to efficiently find optimal paths while significantly reducing search space.
Nevertheless, selecting an admissible heuristic remains challenging due to complex and unknown environments, large search space, and coarse cost map initialization.
In contrast, deep heuristic learning approaches
~\cite{neural_a_star,trans_path,wastar,deep_heuristic}
estimate state priors and cost-to-go values from the current state to the target given expert demonstrations, reducing the search space and enhancing its generalization and adaptability to unseen environments.

Recent advancements have demonstrated significant advantages of using convolutional neural network (CNN)~\cite{cnn,cnn_2,neural_a_star}
and Transformer~\cite{transformer,transformer_2,trans_path}
to learn high-confidence and readily accessible heuristics.
They use the pathfinding algorithm as a post-processing step or in end-to-end supervised learning to learn the heuristic probability map.
However, the post-processing approach heavily relies on an accurate path probability map (PPM) predicted by a neural network, requiring computationally expensive high-confidence PPMs as reference~\cite{trans_path}.
In contrast, end-to-end learning uses only binary labelling to obtain reference path, which is more accessible than obtaining reference PPM.

Among end-to-end differentiable path planning techniques~\cite{gt_path_dijkstra,implicit_bi_opt,bbox,transformer_2,graphcnn},
neural \astar~\cite{neural_a_star} uses \astar~\cite{astar} to learn cost map and discretized activation~\cite{binary_net} to prevent gradient accumulation in path search.
Nevertheless, the balance between state and progressive priors is often imperfect and adaptive edge-wise interactions are rarely exploited, giving inferior path imitation ability.
While these methods aim to find the shortest path with few search steps, they are also inferior in achieving high path similarity to particularly handcrafted trajectories that are not necessarily the shortest for empirical motions in the real world.

To address these challenges, we propose DA\astar~by incorporating distance heuristic and PAF into \astar~for path imitation learning through path shortening and smoothing.
Specifically, PAF is designed to find the best trade-off between the minimum and maximum path angles, acknowledging that the reference path can be arbitrarily smooth and not necessarily the shortest.
We learn PAF by joint learning unary and higher-order priors in a best-first search manner, simultaneous with PPM learning to reduce search space.

To evaluate the effectiveness and generalization of our method, we use 7 datasets, including binary maze, video-game, and drone-view maps.
We demonstrate the superiority of DA\astar~in enhancing path imitation during learning, ensuring that the predicted path is closely aligned with the reference while maintaining a high path shortening rate when applicable.
Furthermore, our DA\astar~outperforms the baselines and state-of-the-art TransPath~\cite{trans_path}, achieving overall higher path optimality with considerably small cost of search efficiency.
Our contributions are summarised as
\begin{itemize}
    \item We investigate the inferiority of the popular \astar~and neural \astar~in imitating reference path which has arbitrary path smoothness and is not necessarily the shortest.
    Our insight and comprehensive study underscore the importance of learning path smoothness to improve the imitation capability from expert demonstrations.
    \item  We propose PAF with min-max path angle optimization to adapt path smoothness in each search step.
    Unlike conventional any-angle algorithms that require every two waypoints at obstacle corners to form linear path segments, we directly impose path angle in the search step.
    This greatly reduces search space and improves its generalization to real scenarios without explicitly finding obstacle corners.
    To the best of our knowledge, this work is the first to comprehensively validate the effect of path angle on deep path imitation learning.
    \item The effectiveness of our DA\astar~in joint learning with and without PPM loss highlights the significance of learning with binary reference paths.
    It notably surpasses the state-of-the-art TransPath that uses computationally expensive PPM as a reference for supervised learning.
\end{itemize}

\section{Related Work}

\paragraph{Conventional Path Planning.}
Conventional pathfinding encompasses search-based and sampling-based algorithms.
Search-based algorithms include Dijkstra's algorithm~\cite{dijkstra}, \astar~\cite{astar}, best-first search~\cite{best_first}, and breadth-first search~\cite{breadth_first}.
Sampling-based algorithms, such as rapidly exploring random trees~\cite{rrt}, probabilistic roadmap~\cite{prob_roadmap}, fast marching trees~\cite{fast_marching}, and their variants~\cite{rrt_connect,anytime_rrt} are also prominent.

Sampling-based algorithms are typically employed in high-dimensional and dynamic environments by constructing probabilistic roadmaps or trees.
In contrast, search-based algorithms operate with a discrete graph representation, traversing active nodes under specific constraints to ensure path completeness.
Among these, \astar~\cite{astar} serves as a bridge between Dijkstra's algorithm~\cite{dijkstra} and greedy best-first search~\cite{best_first}, employing deterministic heuristic to find the shortest path.
Admissible heuristic estimates the cost from the current node to the target, known as h-value, thereby facilitating efficient node expansion to increase search speed.
However, its search efficiency and path optimality are heavily reliant on the heuristic admissibility.

\paragraph{Deep Path Planning.}
Unlike the deterministic yet imperfectly selected heuristics, learning-based path planning with adaptive priors has garnered attention for its ability to enhance the path imitation learning.
Several studies have focused on learning such heuristics from expert demonstrations using neural networks for 2D planar maps~\cite{sail,bbox,learning_heuristics,implicit_bi_opt,trans_path} and 3D space, particularly with robotic arms~\cite{motion_policy}.
Our work specifically concentrates on pathfinding for image-based maps.

To alleviate the limitations of imperfect heuristic in \astar, neural \astar~\cite{neural_a_star,wastar} differentiates a reformulated objective function with a discretized activation paradigm~\cite{binary_net}, allowing for learning on arbitrary maps.
Despite this advancement, however, discontinuous low-cost areas in cost map necessitate massive searches.
Although primarily aiming to find the shortest path, it may not be applied to imitate handcrafted paths which are not always the shortest.

The state-of-the-art method, TransPath~\cite{trans_path}, enhances the map continuity of PPM by using autoencoder~\cite{autoencoder} and Transformer~\cite{transformer}, followed by focal search~\cite{focal_search} on learned PPM to find the optimal path.
However, this fully-supervised approach requires high-quality PPM as a reference for model training.
This is more costly due to the complexity of obtaining pixel-wise priors compared to binary path labelling.
Also, TransPath uses pathfinding as post-processing, limiting its imitation capability without backpropagating through node priors during training.

Meanwhile, although angular constraints have been employed in pathfinding~\cite{turning_cost,grid_angle_constrained}, 
their adoption and adaptation in end-to-end imitation learning are rarely explored.
Traditional angle-aware pathfinding algorithms find the smoothest path~\cite{grid_angle_constrained} or the true shortest path~\cite{theta_star} by accumulating linear path segments by choosing the angle from an obstacle corner to another that can lead to the shortest Euclidean distance.
However, when learning from expert demonstrations, particularly human-labelled trajectories, the path angle is not always small enough to maintain its smoothness and no distinct obstacle corners are available in cases such as a roundabout.
Under these circumstances, the path smoothness becomes dynamic and agnostic to the admissible heuristic. 
We learn the degree of path smoothness through PAF to enhance path imitation capability using joint optimization of path shortening and smoothing.

\section{Preliminaries}

\subsection{Path Planning as Binary Labelling}
Path planning usually refers to the problem of finding the optimal route that connects the source and target while avoiding collision with known obstacles or being restricted from high-risk areas.
When applying pathfinding to an image with a high cost in each of the impermissible pixels and a low cost in each of the accessible pixels, it aims to find binary labelling, assigned 1 for being included in the route and 0 otherwise, from the source pixel to the target pixel such that the accumulated cost of pixels included in the route is the minimum.
We regard this route as the optimal path and the pixel as a node.

This can be formulated as minimizing an objective function with binary labelling $\mb{x}$ and costs $\boldsymbol{\theta}=f(\mc{I},s,t)$ on a graph~\cite{bbox,implicit_bi_opt}, where $\mc{I}$ is an image, $s$ and $t$ are its source and target nodes, and $f(\cdot)$ can be a deterministic computation (for binary maps) or a neural network.
Consider a graph consisting of $N$ nodes, forming set $\mc{V}$, and $M$ edges, forming set $\mc{E}$.
The objective function can be represented using either nodes or edges.
In the \textit{unary} representation by nodes, the labelling is defined as $\mb{x}=\{x_i\}$ and $\boldsymbol{\theta}=\{\theta_i\}$ for all node $i \in \mc{V}$.
Here, the problem size $S=N$ and the labelling constraint function $g(i, \mb{x})=\sum_{j \in \mc{N}_i} x_j$ where $\mc{N}_i$ is the set of $i$'s neighbouring nodes.
Equivalently, for \textit{pairwise} representation by edges, $\mb{x}=\{x_{ij}\}$, $\boldsymbol{\theta}=\{\theta_{ij}\}$ for all edge $(i,j) \in \mc{E}$, the problem size $S=M$, and
the constraint $g(i, \mb{x})=\sum_{j \in \mc{N}_i} x_{ij}$.
The optimal path $\mc{P} \subset \mc{V}$ from $s$ to $t$ is obtained by minimizing the path cost within binary set $\mc{B}^S=\{0, 1\}^S$ as
\begin{equation}
\label{eq:comb_opt}
\begin{aligned}
\min_{\mb{x} \in \mathcal{B}^S} &\boldsymbol{\theta}^{\top} \mb{x}\ ,\\
\text{subject to} \quad g(i, \mb{x}) &=
    \begin{cases}
        2, & \forall i \notin \{s,t\} \\
        1, & \forall i \in \{s,t\} \\
    \end{cases}\ .
\end{aligned}
\end{equation}
This labelling constraint indicates that for a node included in the path if it is the source or target node, only one of its neighbouring nodes is included in the path (\ie only one of its connecting edges forms a path segment); otherwise, two of its neighbouring nodes are included (\ie two of its connecting edges are path segments).

Optimizing Eq.~\eqref{eq:comb_opt} can be computationally intensive for a large-scale problem,
necessitating a reduction of the search space to enhance optimization efficiency.
Therefore, various greedy algorithms have been developed from a dynamic programming perspective~\cite{dijkstra,astar,best_first,breadth_first,rrt,fast_marching}.
Among these, \astar~\cite{astar} is particularly popular due to its guarantee of path completeness and high search efficiency.

\subsection{Heuristic Search}
Unlike breadth-first search~\cite{bfs} and Dijkstra's algorithm~\cite{dijkstra}, which rely solely on the cost-to-start distance (\ie g-value), \astar~\cite{astar} incorporates the cost-to-go distance (\ie h-value) to efficiently reduce search space.
The heuristic distance is calculated on a node-by-node basis.
To be fair, we choose the same heuristics as neural \astar, \ie Chebyshev distance $D^{\text{che.}}_i=\max(\vert p_i - p_t \vert, ~\vert q_i - q_t \vert)$~\cite{cheby} and Euclidean distance $D^{\text{euc.}}_i=\sqrt{(p_i - p_t)^2+(q_i - q_t)^2}$~\cite{euclidean} between Cartesian coordinates $(p_i, q_i)$ of each node $i$ and the target node $t$ for all $i \in \mathcal{V}$.
The weighted heuristic distance is defined as
\begin{equation}
\label{eq:heuristic}
    D_i = D^{\text{che.}}_{i} + \sigma D^{\text{euc.}}_{i}\ ,
\end{equation}
where $D^{\text{che.}}_i$ facilitates search in horizontal and vertical directions while $D^{\text{euc.}}_i$ additionally supports search in diagonal directions. By following the setting in \cite{neural_a_star}, we set $\sigma=0.001$.
Eq.~\eqref{eq:comb_opt} is then modified with this weighted heuristic distance incorporated into the node prior as
\begin{equation}
\begin{aligned}
\label{eq:astar_obj}
    \underset{\mb{x} \in \mc{B}^N} {\min} &\sum_{i \in \mc{V} / \{s,t\}} \left( \theta_i + \lambda D_{i} \right) x_i\ ,\\
    \text{subject to} &\quad \vert| \mc{N}_i \vert|_0 = 2, ~\forall i \in \mc{V} / \{s,t\}\ ,
\end{aligned}
\end{equation}
where $\lambda$ is a weight term to balance the effect of the node prior and the heuristic distance and $\vert| \cdot \vert|_0=2$ indicates that two of $i$'s neighbouring nodes are included in the path.

\begin{remark}
\textit{For a map sharing the same $\theta_i$ for all node $i$ in accessible areas, the heuristic distance, with sufficiently small values, is predominant in finding the shortest path.
For instance, $\theta_i$ for all $i$ in obstacle areas are infinitely large while they are a small constant in accessible areas.
\astar~is often applied to such circumstances, which will be later used to compute path exploration rate (Ep) compared to when $\theta_i$ is no longer a constant for all $i$ in accessible areas.
Note that without the heuristic $D_i$, Eq.~\eqref{eq:comb_opt} is often optimized by using Dijkstra's algorithm which has no heuristic.}
\end{remark}

Therefore, for scenarios without prior knowledge of $\theta_i$, such as video-game and drone-view maps, using \astar~often leads to a direct but incorrect linear path connecting the source node and the target node.
Thus, learning this knowledge from expert demonstrations is crucial for enhancing imitation capability.
This often refers to learning an effective cost map.
In Sec.~\ref{sec:neural_astar}, we will briefly introduce neural \astar~for learning low costs (\ie high probabilities) on accessible areas through the backpropagation of reference path.

\subsection{Neural \astar}
\label{sec:neural_astar}
Neural \astar~integrates CNN and \astar~for end-to-end learning $\boldsymbol{\theta}$ for pathfinding~\cite{neural_a_star}.
Given a reference path, the network learns to assign low costs to areas around the reference path through model training and the dynamic programming of \astar.
The learned node prior map is also known as cost map $\boldsymbol{\theta}$ or path probability map (PPM) which is inversely proportional to $\boldsymbol{\theta}$, meaning that low-cost is equivalent to high-probability.
Moreover, differentiating \astar~involves storing gradients of all intermediate variables derived from $\boldsymbol{\theta}$, leading to high GPU memory consumption due to the numerous search steps.
To address this issue, neural \astar~empirically retains the gradients of $\boldsymbol{\theta}$ while detaching the remainder during path search, significantly reducing GPU memory usage.

\section{Deep Angular \astar}
Although \astar~facilitates the finding of the shortest path using the heuristic guidance in Eq.~\eqref{eq:astar_obj}, its imitation capability is limited by the imperfectly learned cost map.
This is caused by its strong emphasis on path shortening without considering the curvature similarity to the reference path.
Hence, we incorporate smoothness-aware constraints by learning the min-max path angle.
This involves computing the angle between two connecting path segments, followed by learning with the proposed PAF to imitate the path curvature.

\subsection{Smoothness-aware Path Planning}

In conventional path planning, the angular constraint is used to find the smoothest path~\cite{grid_angle_constrained} or the true shortest path~\cite{theta_star}.
They require waypoints at obstacle corners for changing the motion direction, which however is not applicable to scenarios such as roundabouts or without prior obstacle knowledge.
Hence, we impose the angular constraint in every search step over the nodes.
This also benefits the learning of more smooth and effective cost maps.

\paragraph{Path Angular Constraint.}
Given an edge connecting nodes $i$ and $j$ as $e_{ij}=(p_i - p_j,~q_i - q_j)$ where $p$ and $q$ are their 2D Cartesian coordinates, the angle between two connecting edges from node $j$ to node $i$ and then to node $k$ is defined by cosine similarity as
$\theta_{jik} = \arccos \frac{e_{ji}^\top e_{ik}}{\vert| e_{ji} \vert| \vert| e_{ik} \vert|}$.
Then, the minimum path angle can be obtained by
\begin{equation}
\label{eq:angle_opt}
    \min_{\mc{P} \subset \mc{V}} \sum_{i \in \mc{P} / \{s,t\}} \sum_{(j,k) \in \mc{N}_i} \theta_{jik}\ .
\end{equation}
While this aims to find the path with the smallest angle, it is not always preferred.
The path imitation often depends on the similarity of predicted path to reference path considering both path shortening and smoothing for the rationality of actual motions.
Hence, we propose PAF to incorporate the imitation of path smoothness in learning.

\vspace{-3mm}

\paragraph{Path Angular Freedom (PAF).}
Pathfinding guided by the smallest path angle encourages fast search along a linear direction.
However, this may contradict planning practice, where a curved path with a large angle is safer, such as when navigating without collision with obstacles like roundabouts, fences, or lakes, requiring larger search space.
Human path labelling is empirical considering the customary object motions in such specific situations.

\begin{definition}
\textit{(PAF) Instead of simply minimizing or maximizing path angle, a combination is applied to general scenarios. Following notations in Eq.~\eqref{eq:angle_opt}, we define PAF as
\begin{equation}
\label{eq:PAF}
    h_{jik} = \alpha \theta_{jik} + (1 - \alpha) (\pi - \theta_{jik})\ ,
\end{equation}
where $\alpha\in[0, 1]$ is a trade-off between min-max degrees.}
\end{definition}
Similar to Eq.~\eqref{eq:angle_opt}, minimizing $\theta_{jik}$ aims to decrease the path angle, whereas minimizing $(\pi - \theta_{jik})$ increases the path angle.
This facilitates flexibility and automation in adapting the path curvature, yielding more smooth and effective cost maps.
Since $\alpha$ may vary on different datasets, it is challenging to predetermine a constant for all datasets.
In the experiments, we set $\alpha=1$ for DA\astar-min, $\alpha=0$ for DA\astar-max, and use a learned $\alpha$ for DA\astar-mix.
Below, we will explore its impact on end-to-end imitation learning.

\subsection{Joint Optimization with Higher-order Priors}

To this end, we have introduced node prior $\theta_i$, heuristic distance $D_i$ for path shortening, and PAF $h_{jik}$ for path smoothing for all $i \in \mc{V}$ and $(j,k) \in \mc{N}_i$.
Following the unary representation in Eq.~\eqref{eq:astar_obj}, $h_{jik}$ requires nodes $j$, $i$, and $k$ to be included in the path, giving binary label $x_j=x_i=x_k=1$.
This is represented as a higher-order term $h_{jik} x_j x_i x_k$.
The objective function becomes
\begin{align}
\label{eq:joint}
    \underset{\mb{x} \in \mc{B}^N} {\min}\sum_{i \in \mc{V} / \{s,t\}} &\left( \theta_i + \lambda D_{i} \right) x_i + \beta \sum_{(j,k) \in \mc{N}_i} h_{jik} x_j x_i x_k\ ,\nonumber\\
    \text{subject to} &\quad \vert| \mc{N}_i \vert|_0 = 2, ~\forall i \in \mc{V} / \{s,t\}\ ,
\end{align}
where $\lambda, \beta \in [0, 1]$ are learned weights for path shortening and smoothing, respectively.
Since solving Eq.~\eqref{eq:joint} can be computationally expensive, particularly given a large problem size $N$,
greedy algorithms, typically message-passing style~\cite{mplayers,trws}, are advantageous for reducing the problem complexity to enhance search efficiency.

\paragraph{Message-passing Perspective for Node Expansion.}
To perform Eq.~\eqref{eq:joint} from source $s$ to target $t$, sequential message-passing steps are conducted by pushing and updating node prior and edge-based PAF from $s$ to its neighbouring nodes till reaching $t$.
The message is passed to more and more nodes during the path search.
This is called node expansion.
For the consistency of node search from nodes $j$ to $i$ and then to $k$ in Eq.~\eqref{eq:joint},
we use $k$ to denote the expanded node.
Initially, message $m_s=0$ and $m_k=\infty$ for all $k \in \mc{V}/\{s\}$.
Message from node $i$ to $k$ is updated by
\begin{equation}
\label{eq:msg_update}
m_k \xleftarrow{\text{update}}
\begin{cases}
    \min(m_k,~\theta_s), \quad k \in \mc{N}_s \\
    \min(m_k,~\theta_i + m_i + \kappa h_{jik}), k \in \mc{V}/\{s \cup \mathcal{N}_s\}
\end{cases}
\end{equation}
Then, it retains a portion $\lambda$ of the node prior $\theta_k$ on node $k$ and passes the remaining portion ($1-\lambda$) to its neighbours via message $m_k$, which follows
\begin{equation}
\label{eq:cost_update}
    c_k = \lambda (\theta_k + D_k) + (1 - \lambda) m_k\ .
\end{equation}
Till Eqs.~\eqref{eq:msg_update}-\eqref{eq:cost_update} are executed for all $i \in \mathcal{N}_j$ and $k \in \mc{N}_i / \{j \cup \mathcal{N}_j\}$, the next node $n$ for successor generation is
\begin{equation}
\label{eq:softmax}
\begin{aligned}
    n = \argmax_{k \in \mc{O}}~p_k\ ,\quad
    \text{where}~
    p_k = \frac{\exp(-c_k)}{\sum_{k \in \mc{O}} \exp(-c_k)}\ ,
\end{aligned}
\end{equation}
$\mc{O}$ is the open list of active nodes to be traversed, excluding those already traversed,
and $p_k$ for all $k \in \mc{O}$ will then be used to compute the $L_1$ path loss during model training.
The pathfinding terminates when $n = t$, ensuring path completeness.
Here, $\beta= (1 - \lambda) \kappa$ aligns Eq.~\eqref{eq:cost_update} with Eq.~\eqref{eq:joint}.
We provide the pseudo-code of the pathfinding in DA\astar~in Alg.~\ref{alg:daa_code} and its model training in Alg.~\ref{alg:daa_train} of Appendix~\ref{sec:tranining}.

{\small
\begin{algorithm}[t]
\caption{Pathfinding in Deep Angular \astar}
\label{alg:daa_code}
\textbf{Input}: Image $\mc{I}$ with $N$ nodes, source node $s$, and target $t$.\\
\textbf{Parameter}: Learned weights $(\alpha, \lambda, \kappa)$, a trained network $f(\mc{I},s,t)$ to estimate path cost map $\boldsymbol{\theta}$.\\
\textbf{Output}: Optimal path $\mc{P}$.

\begin{algorithmic}[1]
\STATE \textbf{Step 1: Message Passing for Node Expansion}
\STATE Define closed list $\mc{M}_c = \boldsymbol{\emptyset}$, open list $\mc{M}_o = \{s\}$, parent set $\mc{M}_p = \boldsymbol{\emptyset}$, initial message $m_i=\infty$ for all $ i \in \mc{V}$.
\STATE Given target node $t$, compute heuristic distance ${D}_i$ for all $i\in \mc{V}$ by Eq.~\eqref{eq:heuristic}.
\STATE Estimate cost map $\boldsymbol{\theta} = f(\mc{I}, s, t)$.
\STATE Set the current node $i=s$.
\WHILE{\textup{$\mc{M}_o \neq \boldsymbol{\emptyset}$}}
  \STATE Find neighbouring nodes of $i$ as $\mc{N}(i)$.
  \STATE Include $i$ to closed list $\mc{M}_c \gets \mc{M}_c \cup \{i\}$.
  \STATE Include neighbouring nodes to and exclude $i$ from open list $\mc{M}_o \gets \mc{M}_o \cup \mc{N}(i) / \mc{M}_c$.
  \FOR {\textup{$k \in \mc{N}(i)$}}
    \STATE Compute PAF if the angle exists by Eq.~\eqref{eq:PAF}.
    \STATE Update message $m_k$ by Eq.~\eqref{eq:msg_update}.
    \IF {$m_k$ is decreased}
      \STATE Update the parent of $k$ as $i$ by $\mc{M}_p(k) = i$.
    \ENDIF
    \STATE Compute cost $c_k$ by Eq.~\eqref{eq:cost_update}. \\
    \IF {\textup{$k=t$}}
      \STATE break
    \ENDIF
  \ENDFOR
  \STATE Use $c_k$ for all $k \in \mc{M}_o$ to select the next traverse node $i$ for successor generation by Eq.~\eqref{eq:softmax}.
\ENDWHILE
\STATE \textbf{Step 2: Backtracking for Optimal Path}
\STATE Set the current node $i=t$ and $\mc{P} = \{t\}$.\\
\WHILE{\textup{$i \neq s$}}
  \STATE Find the parent node of $i$ by $\mc{M}_p(i)$.\\
  \STATE Include the parent node to the path $\mc{P} \gets \mc{P} \cup \{i\}$.\\
  \STATE Update the current node $i$ by its parent node $\mc{M}_p(i)$.\\
\ENDWHILE
\STATE \textbf{return} $\mc{P}$\\
\end{algorithmic}
\end{algorithm}
}

\section{Experiments}

We refer to DA\astar-mix as DA\astar~by default.
All GPU-related experiments are conducted on one RTX 3090.
Code is available at \href{https://github.com/zwxu064/DAAStar.git}{https://github.com/zwxu064/DAAStar.git}.

\subsection{Setup}
\label{sec:setup}

\paragraph{Datasets.}
To analyse the effectiveness and generalization of our method, we evaluate it on 7 datasets with a relatively wide range of map types, reference path types, and data augmentations.
See an overview of the datasets in Table~\ref{tb:dataset}.
%
\begin{table}[t]
\centering
\setlength{\tabcolsep}{4pt}
\resizebox{0.47\textwidth}{!}{
\begin{tabular}{l|rrrc}
\Xhline{2\arrayrulewidth}
& \multicolumn{1}{c}{\multirow{1}{*}{\textbf{Train}}} & \multicolumn{1}{c}{\textbf{Val}} & \multicolumn{1}{c}{\multirow{1}{*}{\textbf{Test}}} & \multicolumn{1}{c}{\multirow{1}{*}{\textbf{Graph}}} \\
\Xhline{2\arrayrulewidth}
\multicolumn{1}{l|}{MPD} & 800 & 100 ($\times$6) & 100 ($\times$15) & 32$\times$32 \\
\multicolumn{1}{l|}{TMPD} & 3,200 & 400 ($\times$6) & 400 ($\times$15) & 64$\times$64 \\
\multicolumn{1}{l|}{CSM} & 3,200 & 400 ($\times$6) & 400 ($\times$15) & 64$\times$64 \\
\multicolumn{1}{l|}{Warcraft} & 2,500 ($\times$4) & 250 ($\times$4) & 250 ($\times$4) & 12$\times$12 \\
\multicolumn{1}{l|}{Pok\'emon} & 750 ($\times$4) & 125 ($\times$4) & 125 ($\times$4) & 20$\times$20 \\
\multirow{1}{*}{SDD-intra} & 6,847 & 1,478 & 1,478 & 64$\times$64 \\
\multirow{1}{*}{SDD-inter} & 7,284 & 1,040 & 1,040 & 64$\times$64 \\
Aug-TMPD & 512,000 & 64,000 & 64,000 & 64$\times$64 \\
\Xhline{2\arrayrulewidth}
\end{tabular}
}
\caption{Dataset overview.
The amount of MPD maps is on each scene and the one of SDD-inter is averaged on 8 scenes.}
\label{tb:dataset}
\vspace{-5mm}
\end{table}

\begin{itemize}[leftmargin=*]
    \item \textbf{Maze Datasets}. We evaluate DA\astar~on 3 datasets:
    {motion planning dataset (MPD)} with 8 scenes~\cite{mp},
    {tiled motion planning dataset (TMPD)}~\cite{mp}, and city and street maps (CSM)~\cite{city_street}.
    Following MPD, we sample 15 random source nodes, 5 from each of 3 regions in the 55\%--70\%, 70\%--85\%, and 85\%--100\% percentile points.
    Reference paths are computed using optimal policies provided by~\cite{neural_a_star}, which gives rich motion turning and achieves shorter path length than \astar~and \tstar, evidenced by shortest path ratio$<$100\% in Tables~\ref{tb:full_table_maze}--\ref{tb:sota}.
    \item \textbf{Video-game Datasets}.
    Warcraft and Pok\'emon are used with 4$\times$ data augmentations~\cite{wastar}.
    Each map contains 2 target nodes while each with 2 source nodes.
    The image size of Warcraft is 96$\times$96, downsampled by 1/8 as graph size, and 320$\times$320, downsampled by 1/16, for Pok\'emon.
    The reference paths are computed by Dijkstra's algorithm on given reference cost maps, see Figs.~\ref{fig:examples}(\iccvcolor{c})--(\iccvcolor{d}).
    
    \item \textbf{Drone-view Dataset}. The Stanford drone dataset (SDD)~\cite{sdd} contains 8 scenes with human-labelled paths, containing intra-scene and inter-scene scenarios~\cite{sdd_2}.
    Following the setting in~\cite{neural_a_star}, for the intra-scene experiments, we use one video from each scene for testing and the rest for training; 
    for the inter-scene experiments, we use one scene for testing and the other 7 scenes for training.

    \item \textbf{Augmented TMPD.} The state-of-the-art TransPath learns PPM with reference PPM on augmented TMPD (Aug-TMPD) in a fully-supervised fashion~\cite{trans_path}.
    As post-processing, it uses focal search~\cite{focal_search} to find optimal paths within high-probability areas of PPM predicted by Transformer.
    The reference PPM can be computed as the search history of \tstar~\cite{theta_star}.
    To compare supervised learning with the $L_1$ path loss, we compute reference paths using Dijkstra's algorithm on the binary maze maps.
\end{itemize}

\vspace{-5mm}
\paragraph{Evaluation Metrics.}

Different from those regarding the shortest path as the optimal, human-labelled paths on SDD are not the shortest.
Hence, we refer to path optimality as path similarity between the prediction and reference paths.
We also consider the shortest path to be optimal if the reference path is the shortest.
Therefore, we use the path optimality ratio (Opt) from~\cite{bbox} as the shortest path ratio (SPR) to measure the \textit{path shortening rate}.
The \textit{path similarity} is measured by path alignment similarity (PSIM), area similarity (ASIM), and Chamfer distance (CD)~\cite{cdist}, between the prediction and reference.
When the predicted path is identical to the reference, they are fully aligned without any areas in between, giving full PSIM and ASIM.
The \textit{search efficiency} is measured by search history (Hist) and the reduction ratio of path exploration (Ep)~\cite{neural_a_star} where \astar~is applicable.
Hist is the ratio of traversed nodes to image size and Ep is the reduction of Hist over \astar.
Given $K$ samples indexed by $i$, the metrics are
{\small
\begin{align}
\text{SPR:\quad }&
\frac{1}{K} \sum_{i=1}^K \mathbb{1} \left[ \vert| \mc{P}_i \vert|_0 \leq \vert| \hat{\mc{P}}_i \vert|_0 \right]\ ,\nonumber\\
\text{PSIM:\quad}&
1 - \frac{1}{K} \sum_{i=1}^K \min (\frac{\vert| \mc{P}_i - \hat{\mc{P}}_i \vert|_0}{2~\vert| \hat{\mc{P}}_i \vert|_0}, 1) \ ,\nonumber\\
\text{ASIM:\quad}&
1 - \frac{1}{K} \sum_{i=1}^K \frac{\texttt{Area}(\mc{P}_{ij}, \hat{\mc{P}}_i)}{\cup_{j \in \mc{J}} \texttt{Area}(\mc{P}_{ij}, \hat{\mc{P}}_i)}\ ,\nonumber\\
\text{CD:~}
\frac{1}{K} \sum_{i=1}^K ( \sum_{{x} \in \mc{I}_i} &\min_{{y} \in \hat{\mc{I}}_i} \vert| {x} - {y} \vert|_2^2 + \sum_{{y} \in \hat{\mc{I}}_i} \min_{{x} \in \mc{I}_i} \vert| {x} - {y} \vert|_2^2 )\ ,\nonumber\\
\text{Hist:\quad}&
\frac{1}{K} \sum_{i=1}^K \frac{\vert| \mc{M}^c_i \vert|_0}{H W}
\ ,\nonumber \\
\text{Ep:\quad}&
\frac{1}{K} \sum_{i=1}^K \frac{\max(\vert| \hat{\mc{M}}^c_i \vert|_0 - \vert| \mc{M}^c_i \vert|_0, 0)}{\vert| \hat{\mc{M}}^c_i \vert|_0}\ ,\nonumber
\end{align}
}
where ${\mc{P}}_i$ is the predicted path, $\mc{P}_{ij}$ is the predicted path by $j$ method, $\hat{\mc{P}}_i$ is the reference path, the function $\texttt{Area}(\mb{a}, \mb{b})$ computes the area between two discrete paths $\mb{a}$ and $\mb{b}$ \cite{area}, $H$ and $W$ are image height and width, $\hat{\mc{M}}^c_i$ is the closed list (containing all traversed nodes) obtained by \astar, ${\mc{M}}^c_i$ is the closed list obtained by a selected pathfinding method,
and $\hat{\mc{I}}_i$ and ${\mc{I}}_i$ are 2D coordinates of $\hat{\mc{P}}_i$ and $\mc{P}_i$, respectively.

\vspace{-2mm}

\paragraph{Hyper-parameters for Model Training.}
To reproduce the experiments, we provide main hyper-parameters for model training.
For \textit{the maze datasets}, we use UNet as the cost map encoder and train it for 400 epochs with batch size 100 and learning rate $1\mathrm{e}^{-3}$.
For \textit{the video-game datasets}, the cost map encoder is CNN with $M$ block(s) of ``Conv2D--BatchNorm2D--ReLU--MaxPool2D," downsampled by half with each output channel 32$\times$2$^{M-1}$, followed by a 2D convolutional layer and Sigmoid activation, where $M=3$ for Warcraft and $M=4$ for Pok\'emon.
We train the models for 100 epochs with batch size 64 and learning rate $5\mathrm{e}^{-4}$.
For \textit{the drone-view dataset}, we train UNet for 50 epochs on inter-scene and 150 epochs on intra-scene with batch size 64 and learning rate $1\mathrm{e}^{-3}$.
$\boldsymbol{\theta}$ is scaled by 10 after Sigmoid activation.
We train these models with 3 random seeds and report the mean and standard deviation.
For \textit{Aug-TMPD}, we use Transformer as PPM encoder and train it for 50 epochs with batch size 64 and learning rate $1\mathrm{e}^{-3}$.

\subsection{Comparison Methods}
\label{sec:comparison}

\paragraph{\astar~and \tstar.}
\astar~and \tstar~are typical traditional pathfinding methods without and with the consideration of search angle.
Particularly, \astar~facilitates the search steps over Dijkstra's algorithm with heuristic guidance from the shortest distance between the source and target nodes.
\tstar~finds the true shortest path by accumulating the shortest Euclidean distance between each two waypoints at obstacle corners~\cite{theta_star}.
Similar to \astar, it requires finding the obstacles and the nodes of path, which can be computed by Bresenham's algorithm~\cite{Bresenham} as post-processing.
However, since such traditional methods require prior knowledge of obstacles, we only apply them to binary maps that indicate the obstacles.
This avoids predicting linear paths that are incorrect by simply connecting the source and target nodes.

\vspace{-5mm}

\paragraph{Neural \astar.}
The main deep learning baseline is neural \astar~\cite{neural_a_star}.
It enables the learning ability of \astar~\cite{astar} to predict effective cost maps to reduce search space.
While \astar~generally achieves a much higher SPR than neural \astar~because of its high search history (Hist), its path similarity to the reference is much lower.
This is also applied to \tstar.
Neural \astar~computes the $L_1$ loss between the predicted and reference path, yielding effective cost maps. 

\vspace{-5mm}

\paragraph{Random-walk Search.}
Incorporating path angular constraint into pathfinding perturbs the node expansion in accessible areas, alleviating the effect of heuristic distance.
Given path completeness, we randomly shuffle node expansion within top-$k$ least-cost nodes with probability normalized by Eq.~\eqref{eq:softmax}.
This alters the traverse order and possibly the search direction, resulting in better path imitation than neural \astar, particularly on SDD shown in Table~\ref{tb:full_table_sdd}.
We examine $k \in \{3,5\}$ and demonstrate $k=3$ as a competitive comparison considering its higher path optimality.

\vspace{-5mm}
\paragraph{State-of-the-art TransPath.}
TransPath~\cite{trans_path} benefits path search from continuous PPM predicted by Transformer due to the multi-head self-attention features.
It provides supervised learning on reference PPM, where the area with values greater than threshold 0.95 is the closed list $\mc{M}^c$ of \tstar, which can be verified using motion planning library\footnotemark.
During training, the Transformer learns node priors with reference PPM computed by \tstar.
During testing, it uses focal search to find the path on the predicted PPM as post-processing.
We apply PPM loss only on Aug-TMPD given the reference PPMs from~\cite{trans_path}.

\footnotetext{Source code: \href{https://github.com/ai-winter/python_motion_planning}{https://github.com/ai-winter/python\_motion\_planning}.}

\subsection{Main Results and Analysis}

\paragraph{Evaluation on Maze Datasets.}
In binary maze maps, accessible areas share constant node priors, resulting in non-unique solutions of the shortest path.
However, to learn from reference path, the prediction should maintain a high shortening rate while closely aligning with the reference.

\begin{figure*}[!ht]
\centering
\includegraphics[width=\textwidth]{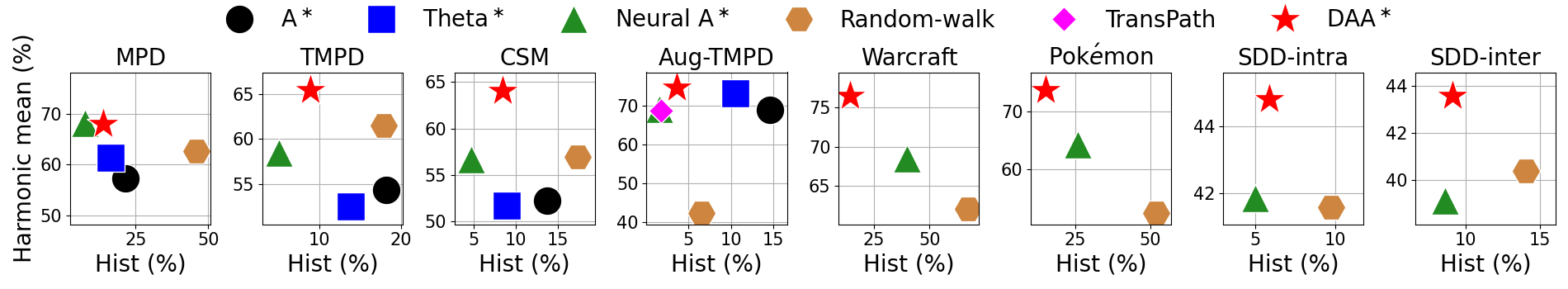}
\vspace{-7mm}
\caption{Trade-off between search efficiency and path optimality measured by Harmonic mean (HMean): SIM=HMean(PSIM,ASIM) or HMean(SPR, SIM) if SPR is available in Tables~\ref{tb:full_table_maze}--\ref{tb:full_table_sdd}.
DA\astar~refers to DA\astar-weight on Aug-TMPD and DA\astar-mix on the others.
}
\label{fig:trade-off}
\vspace{-4mm}
\end{figure*}

In Table~\ref{tb:full_table_maze}, DA\astar~achieves the highest SPR among all learning-based methods, the highest path similarity, and largely reduced search history over \astar~across all datasets.
While \astar~and \tstar~achieve higher SPR, they require more search steps, shown in Fig.~\ref{fig:trade-off}.
The learned $\alpha$ of DA\astar-mix for the datasets is 0.33, 0.77, and 0.69 accordingly.

\begin{table}[t]
\centering
\setlength{\tabcolsep}{3pt}
\resizebox{0.47\textwidth}{!}{
\begin{tabular}{l|cccc}
\Xhline{2\arrayrulewidth}
\multicolumn{1}{c|}{\multirow{1}{*}{\textbf{Method}}} & \multicolumn{1}{c}{\textbf{SPR} (\%)$\uparrow$} & \multicolumn{1}{c}{\textbf{PSIM} (\%)$\uparrow$} & \multicolumn{1}{c}{\textbf{ASIM} (\%)$\uparrow$} & \multicolumn{1}{c}{\textbf{Ep} (\%)$\uparrow$} \\
\Xhline{2\arrayrulewidth}
\multicolumn{5}{c}{MPD} \\
\hline
\astar & \bestcolor{\acc{98.7}{0.0}} & \acc{35.2}{0.0} & \acc{47.27}{0.0} & N/A \\
\tstar & \bestcolor{\acc{98.7}{0.0}} & \acc{38.0}{0.0} & \acc{53.84}{0.0} & N/A \\
\hdashline
Neural \astar & \acc{91.19}{0.37} & \acc{44.26}{0.20} & \acc{54.93}{0.10} & \acc{44.10}{0.26} \\
Rand-walk & \acc{82.06}{1.11} & \acc{46.63}{0.15} & \acc{55.72}{0.07} & \hspace{0.8mm}
\acc{0.18}{0.01}
\\
DA\astar-min & \acc{91.56}{0.14} & \acc{45.22}{0.28} & \acc{55.53}{0.06} & \acc{53.43}{0.36} \\
DA\astar-max & \acc{93.53}{0.32} & \bestcolor{\acc{47.95}{0.15}} & \bestcolor{\acc{58.89}{0.08}} & \bestcolor{\acc{70.96}{0.42}} \\
DA\astar-mix & \secbestcolor{\acc{95.56}{0.42}} & \secbestcolor{\acc{47.83}{0.14}} & \secbestcolor{\acc{58.72}{0.12}} & \secbestcolor{\acc{69.99}{0.09}} \\
\Xhline{2\arrayrulewidth}
\multicolumn{5}{c}{TMPD} \\
\hline
\astar & \bestcolor{\acc{94.7}{0.0}} & \acc{29.4}{0.0} & \acc{54.07}{0.0} & N/A \\
\tstar & \secbestcolor{\acc{91.8}{0.0}} & \acc{28.6}{0.0} & \acc{51.42}{0.0} & N/A \\
\hdashline
Neural \astar & \acc{78.63}{0.56} & \acc{39.04}{0.10} & \acc{57.47}{0.06} & \acc{49.86}{1.12} \\
Rand-walk & \acc{78.30}{1.12} & \secbestcolor{\acc{43.08}{0.61}} & \secbestcolor{\acc{61.16}{0.11}} & \hspace{1.8mm}\acc{9.51}{0.44} \\
DA\astar-min & \acc{81.91}{1.39} & \acc{40.37}{0.29} & \acc{59.03}{0.18} & \acc{63.63}{0.39} \\
DA\astar-max & \acc{80.13}{0.38} & \acc{40.42}{0.30} & \acc{58.93}{0.06} & \bestcolor{\acc{84.03}{0.67}} \\
DA\astar-mix & \acc{88.59}{0.30} & \bestcolor{\acc{43.86}{0.44}}
& \bestcolor{\acc{63.29}{0.03}}
& \secbestcolor{\acc{78.90}{1.42}} \\
\Xhline{2\arrayrulewidth}
\multicolumn{5}{c}{CSM} \\
\hline
\astar & \bestcolor{\acc{94.6}{0.0}} & \acc{27.2}{0.0} & \acc{53.61}{0.0} & N/A \\
\tstar & \secbestcolor{\acc{93.67}{0.0}} & \acc{27.3}{0.0} & \acc{51.53}{0.0} & N/A \\
\hdashline
Neural \astar & \acc{73.83}{0.06} & \acc{38.64}{0.21} & \acc{56.85}{0.10} & \acc{30.92}{0.52} \\
Rand-walk & \acc{70.38}{1.06} & \secbestcolor{\acc{43.77}{0.15}} & \secbestcolor{\acc{62.67}{0.05}} & \hspace{1.8mm}\acc{2.57}{0.10} \\
DA\astar-min & \acc{80.16}{1.32} & \acc{42.96}{0.73} & \acc{62.14}{0.08} & \acc{44.70}{1.90} \\
DA\astar-max & \acc{74.08}{1.45} & \acc{40.73}{0.55} & \acc{59.28}{0.14} & \bestcolor{\acc{83.33}{1.20}} \\
DA\astar-mix & \acc{82.03}{1.13} & \bestcolor{\acc{44.51}{0.76}} & \bestcolor{\acc{64.15}{0.08}} & \secbestcolor{\acc{76.86}{1.31}} \\
\Xhline{2\arrayrulewidth}
\end{tabular}
}
\caption{Path optimality on maze datasets. We highlight learning-based methods due to their high search efficiency measured by Ep and Hist in Fig.~\ref{fig:trade-off} and provide examples in Fig.~\ref{fig:examples}(\iccvcolor{a}).}
\label{tb:full_table_maze}
\vspace{-5mm}
\end{table}

In Table~\ref{tb:sota}, DA\astar~trained with PPM loss and path loss, by both DA\astar-mask ($\alpha=0.7$) and DA\astar-weight ($\alpha=0.62$), or even with only path loss by DA\astar-path ($\alpha=0.59$) significantly outperform the baselines and TransPath with shorter length, higher similarity, and fewer search steps.
Particularly, DA\astar-mask filters out areas distant from reference path to alleviate the side-effect of reference PPM on path imitation because low-cost PPM areas not included in the reference path will increase path loss during training.
DA\astar-weight uses a higher weight, 10$\times$, on path loss than PPM loss to emphasize the importance of reference path.
One should also note that it is often easier to obtain binary paths than PPMs for supervised learning, especially for real-world maps.
Hence, training using path loss as in neural \astar~and DA\astar~is more practical than TransPath using PPM.
The learned $\lambda$ and $\kappa$ are provided in Appendix~\ref{sec:weights}.

\begin{table}[t]
\centering
\setlength{\tabcolsep}{8pt}
\resizebox{0.46\textwidth}{!}{
\begin{tabular}{l|ccc|c}
\Xhline{2\arrayrulewidth}
\multicolumn{1}{c|}{\textbf{Method}} & \textbf{SPR}$\uparrow$ & \textbf{PSIM}$\uparrow$ & \textbf{ASIM}$\uparrow$ & \textbf{Hist}$\downarrow$ \\
\Xhline{2\arrayrulewidth}
\astar & \secbestcolor{99.08} & 52.61 & 52.96 & \hspace{-1.8mm}14.59 \\
\tstar & \bestcolor{99.65} & 51.76 & 65.18 & \hspace{-1.8mm}10.53 \\
\hdashline
Neural \astar & 90.92 & 50.61 & 62.11 & \bestcolor{1.59} \\
Rand-walk & 37.22 & 45.46 & 54.31 & 6.57 \\
TransPath\footnotemark & 90.62 & 49.78 & 62.53 & 1.83 \\
DA\astar & 87.04 & 53.38 & 62.14 & \secbestcolor{1.68} \\
DA\astar-path & 94.23 & \bestcolor{56.37} & 65.44 & 4.02 \\
DA\astar-mask & 96.04 & 54.91 & \secbestcolor{65.87} & 4.02 \\
DA\astar-weight & 96.87 & \secbestcolor{55.73} & \bestcolor{66.20} & 3.66 \\
\Xhline{2\arrayrulewidth}
\end{tabular}
}
\caption{Joint learning with and w/o PPM loss on Aug-TMPD.
 We highlight learning-based methods for high search efficiency with examples in Fig.~\ref{fig:examples}(\iccvcolor{b}). DA\astar~refers to DA\astar-mix with both PPM loss and path loss, and DA\astar-path uses DA\astar-mix with path loss.}
\label{tb:sota}
\vspace{-2mm}
\end{table}

\footnotetext{We use weights from \href{https://github.com/AIRI-Institute/TransPath.git}{https://github.com/AIRI-Institute/TransPath.git}.}

\begin{table}[t]
\centering
\setlength{\tabcolsep}{2.5pt}
\resizebox{0.47\textwidth}{!}{
\begin{tabular}{l|cccc}
\Xhline{2\arrayrulewidth}
\multicolumn{1}{c|}{\multirow{1}{*}{\textbf{Method}}} & \multicolumn{1}{c}{\textbf{SPR} (\%)$\uparrow$} & \multicolumn{1}{c}{\textbf{PSIM} (\%)$\uparrow$} & \multicolumn{1}{c}{\textbf{ASIM} (\%)$\uparrow$} & \multicolumn{1}{c}{\textbf{CD}$\downarrow$} \\
\Xhline{2\arrayrulewidth}
\multicolumn{5}{c}{Warcraft} \\
\hline
Neural \astar & \acc{81.03}{0.58} &
\acc{70.65}{0.60} & \acc{51.05}{0.33} & \hspace{1.8mm}\acc{2.85}{0.15} \\
Rand-walk & \acc{72.68}{0.97} & \acc{67.95}{0.40} & \acc{45.09}{0.11} & \hspace{1.8mm}\acc{2.94}{0.04}  \\
DA\astar-min & \acc{86.55}{1.26} & \bestcolor{\acc{76.71}{0.95}} & \bestcolor{\acc{62.68}{0.35}} & \hspace{1.8mm}\bestcolor{\acc{2.29}{0.17}} \\
DA\astar-max & \secbestcolor{\acc{86.83}{0.53}} & \secbestcolor{\acc{75.27}{0.40}} & \acc{60.16}{0.06} & \secbestcolor{\hspace{1.8mm}\acc{2.47}{0.02}} \\
DA\astar-mix & \bestcolor{\acc{89.30}{0.46}} & \acc{75.17}{0.35}
& \secbestcolor{\acc{60.35}{0.01}}
& \hspace{1.8mm}\acc{2.53}{0.02} \\
\Xhline{2\arrayrulewidth}
\multicolumn{5}{c}{Pok\'emon} \\
\hline
Neural \astar & \acc{67.53}{2.80} & \acc{68.84}{1.00} & \acc{55.07}{0.27} & \hspace{1.8mm}\acc{4.94}{0.44} \\
Rand-walk & \acc{52.98}{1.59} & \acc{63.60}{1.24} & \acc{43.77}{0.14} & \hspace{1.8mm}\acc{5.19}{0.26} \\
DA\astar-min & \secbestcolor{\acc{81.25}{2.95}} & \bestcolor{\acc{72.45}{0.79}} & \bestcolor{\acc{61.34}{0.17}} & \hspace{1.8mm}\bestcolor{\acc{4.81}{0.23}} \\
DA\astar-max & \acc{78.00}{2.12} & \acc{71.11}{0.32} & \acc{58.45}{0.03} & \hspace{1.8mm}\secbestcolor{\acc{4.82}{0.09}} \\
DA\astar-mix & \bestcolor{\acc{84.70}{0.76}} & \secbestcolor{\acc{71.49}{0.44}} & \secbestcolor{\acc{60.02}{0.06}} & \hspace{1.8mm}\acc{5.18}{0.17} \\
\Xhline{2\arrayrulewidth}
\end{tabular}
}
\caption{Path optimality on video-game datasets. We only evaluate learning-based methods predicting PPM for effective search in accessible areas.
See examples in Figs.~\ref{fig:examples}(\iccvcolor{c})--(\iccvcolor{d}).}
\label{tb:full_table_game}
\vspace{-5mm}
\end{table}

\begin{figure}[!ht]
\centering
\includegraphics[width=0.48\textwidth]{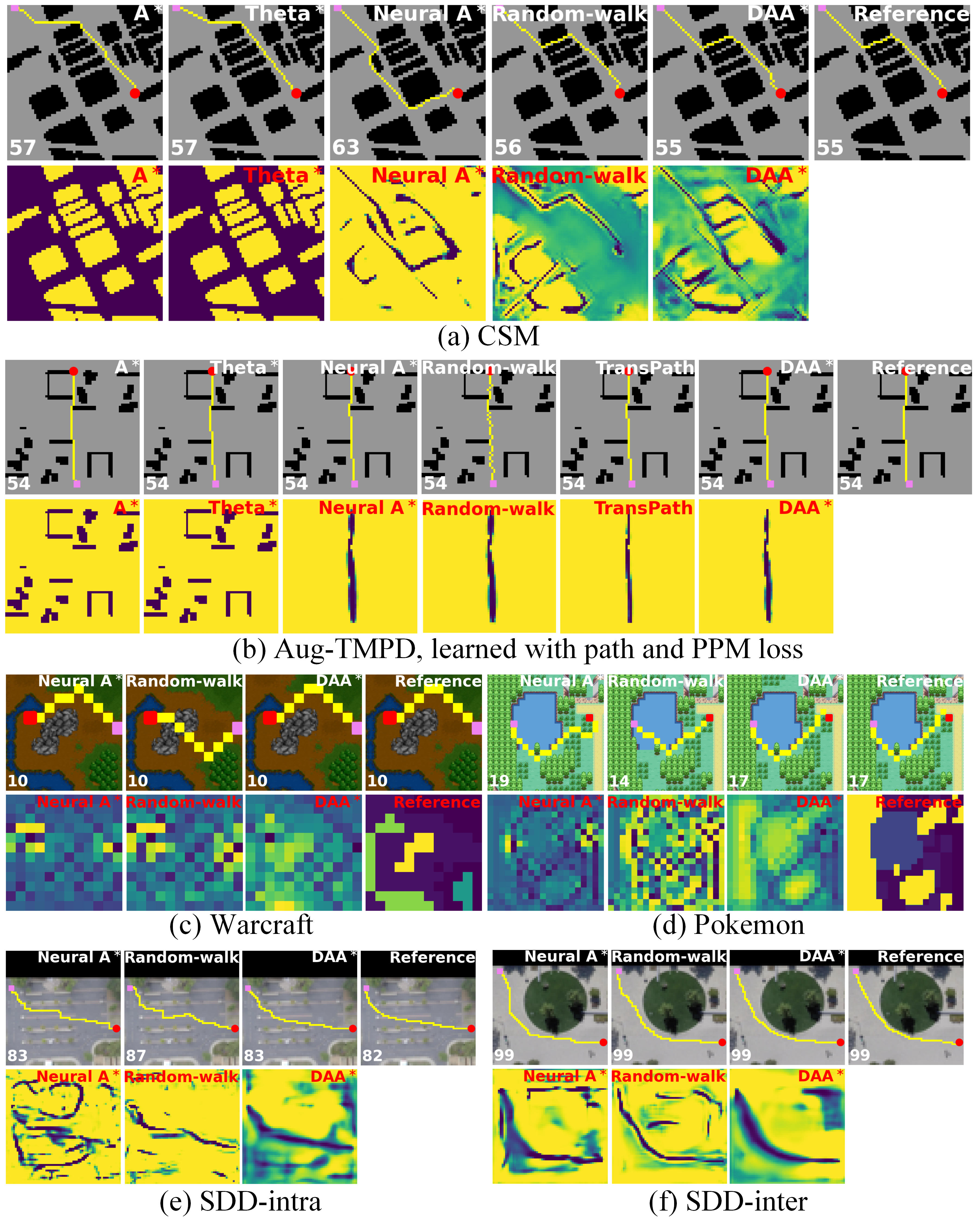}
\vspace{-6mm}
\caption{Examples of optimal path (odd row) and learned cost map with darker colour for lower cost (even row).
Path search is from the red node to the violet node.
Pathfinding method is at the top and path length at the bottom.
We select outstanding methods for visualization, and more examples will be available in our repository along with trained models.
}
\label{fig:examples}
\vspace{-7mm}
\end{figure}

\vspace{-5mm}
\paragraph{Evaluation on Video-game Datasets.}
Unlike maze datasets sharing the same node prior for accessible areas, video-game datasets have diverse impassable areas, including grey stone, blue swamp, and green forest in Fig.~\ref{fig:examples}, requiring different node priors.
In Table~\ref{tb:full_table_game}, our DA\astar~is capable of finding shorter and more similar paths to the reference than the others.
The learned $\alpha$ for Warcraft and Pok\'emon is 0.28 and 0.32, respectively.
Notably, as shown by DA\astar-min's superior performance, a small path angle is favoured.
This may be caused by learning from the reference path and strongly-distinct and densely-clustered objects in the learned PPM, assigning accessible areas with small costs.

\vspace{-5mm}
\paragraph{Evaluation on Drone-view Dataset.}

Given a handcrafted reference path, the path imitation emphasizes both path shortening and smoothing derived from human intuition. Vanilla pathfinding is often challenging to choose realistic path smoothness without analysing the scenario's prior knowledge.
Thus, it is essential to learn PAF to imitate the path smoothness from empirical demonstrations instead of presuming the path angle to be either the minimum or maximum.
The reference path generally exhibits a high path curvature, reflecting the importance of stable motions to avoid traffic accidents in reality.
Meanwhile, only a few scenes encourage linear paths with a small path angle.
Table~\ref{tb:full_table_sdd} shows the similarity between the prediction and reference, measured by PSIM, ASIM, and CD.
Our DA\astar-mix, with $\alpha=0.1$ for SDD-intra and $\alpha=0.17$ for SDD-inter, greatly outperforms the others and over the baseline neural \astar.

\begin{table}[t]
\centering
\setlength{\tabcolsep}{8.5pt}
\resizebox{0.47\textwidth}{!}{
\begin{tabular}{l|ccc}
\Xhline{2\arrayrulewidth}
\multicolumn{1}{c|}{\multirow{1}{*}{\textbf{Method}}} & \multicolumn{1}{c}{\textbf{PSIM} (\%)$\uparrow$} & \multicolumn{1}{c}{\textbf{ASIM} (\%)$\uparrow$} & \multicolumn{1}{c}{\textbf{CD}$\downarrow$} \\
\Xhline{2\arrayrulewidth}
\multicolumn{4}{c}{SDD-intra} \\
\hline
Neural \astar & \acc{40.12}{0.35} & \acc{43.78}{0.04} & \acc{12.25}{0.57} \\
Rand-walk & \acc{40.22}{0.56} & \acc{43.06}{0.02} & \acc{10.42}{0.67} \\
DA\astar-min & \acc{40.44}{0.64} & \acc{44.37}{0.04} & \acc{11.70}{0.61} \\
DA\astar-max & \secbestcolor{\acc{41.90}{0.66}} & \secbestcolor{\acc{47.45}{0.09}} & \hspace{1.8mm}\secbestcolor{\acc{9.39}{1.24}} \\
DA\astar-mix & \bestcolor{\acc{42.12}{0.41}} & \bestcolor{\acc{47.82}{0.02}} & \hspace{1.8mm}\bestcolor{\acc{9.03}{0.16}} \\
\Xhline{2\arrayrulewidth}
\multicolumn{4}{c}{SDD-inter} \\
\hline
Neural \astar & \acc{35.52}{0.05} & \acc{43.52}{0.05} & \acc{23.42}{0.88} \\
Rand-walk & \acc{36.67}{0.28} & \acc{44.96}{0.08} & \acc{19.86}{1.56} \\
DA\astar-min & \acc{35.71}{0.31} & \acc{44.50}{0.02} & \acc{22.47}{0.92} \\
DA\astar-max & \secbestcolor{\acc{38.65}{0.22}} & \secbestcolor{\acc{49.51}{0.09}} & \bestcolor{\acc{18.92}{1.17}} \\
DA\astar-mix & \bestcolor{\acc{38.78}{0.19}} & \bestcolor{\acc{49.77}{0.03}} & \secbestcolor{\acc{19.67}{0.90}} \\
\Xhline{2\arrayrulewidth}
\end{tabular}
}
\caption{Path optimality on SDD. We evaluate learning-based methods predicting effective PPMs, showcased by Figs.~\ref{fig:examples}(\iccvcolor{e})--(\iccvcolor{f}).}
\label{tb:full_table_sdd}
\vspace{-3mm}
\end{table}

\subsection{Trade-off between Optimality and Efficiency}

Conventionally, Dijkstra's algorithm is effective in finding the shortest path using only node priors, but it suffers from low search efficiency.
\astar~and neural \astar~accelerate the search process but have low path similarity in imitation learning, particularly when path smoothness is uncertain.
Fig.~\ref{fig:trade-off} examines the trade-off between path optimality and search efficiency.
Our DA\astar~perturbs the search by considering both path shortening and smoothing, increasing Hist by 0.5\%--6.4\% on the maze datasets and SDD while decreasing it by 10.5\%--25.7\% on the video-game datasets.

\section{Conclusion and Future Work}
We have introduced DA\astar~for pathfinding on 2D images, incorporating PAF to enhance path similarity between the prediction and reference.
Our method adaptively imitates path smoothness by leveraging expert demonstrations in supervised learning.
Extensive experiments on synthetic and real datasets have demonstrated the effectiveness of DA\astar~compared to the baseline and state-of-the-art methods.
Our DA\astar~significantly improves path optimality by learning through path shortening and smoothing with a considerably minor cost of search efficiency.
In video-game datasets, DA\astar~not only improves path optimality but also achieves higher search efficiency.
Even without reference PPM, DA\astar~trained with only path loss can achieve competitive path imitation capability.
In future work, we would consider more complex scenarios, including collision-free navigation and multiple sources with higher search and training efficiency.
These can possibly be achieved by more fine-grain PPM prediction and efficient backpropagation.
We discuss its ethical and societal impact in Appendix~\ref{sec:impact}.

\section*{Acknowledgement}
I sincerely appreciate Prof. Stephen Gould from The Australian National University (ANU) for his invaluable advice, support, and affable smiles all these years and Mr. Jiahao Zhang from ANU for managing and solving GPU-related issues.
If you need my help or support in the future, please do not hesitate to contact me!
I would like to specially thank Prof. Tom Drummond from The University of Melbourne for the financial support of the conference registration.

{
\small
\bibliography{references}

\begin{thebibliography}{60}
\providecommand{\natexlab}[1]{#1}
\providecommand{\url}[1]{\texttt{#1}}
\expandafter\ifx\csname urlstyle\endcsname\relax
  \providecommand{\doi}[1]{doi: #1}\else
  \providecommand{\doi}{doi: \begingroup \urlstyle{rm}\Url}\fi

\bibitem[Alberto~Archetti(2021)]{wastar}
Matteo~Matteucci Alberto~Archetti, Marco~Cannici.
\newblock Neural weighted {A}*: Learning graph costs and heuristics with
  differentiable anytime {A}*.
\newblock In \emph{arXiv:2105.01480}, 2021.

\bibitem[Alesiani(2023)]{implicit_bi_opt}
Francesco Alesiani.
\newblock Implicit bilevel optimization: Differentiating through bilevel
  optimization programming.
\newblock In \emph{AAAI}, 2023.

\bibitem[Amirian et~al.(2020)Amirian, Zhang, Castro, Baldelomar, Hayet, and
  Pettre]{sdd_2}
Javad Amirian, Bingqing Zhang, Francisco~Valente Castro, Juan~Jose Baldelomar,
  Jean-Bernard Hayet, and Julien Pettre.
\newblock {OpenTraj}: Assessing prediction complexity in human trajectories
  datasets.
\newblock In \emph{ACCV}, 2020.

\bibitem[Baldi(2012)]{autoencoder}
Pierre Baldi.
\newblock Autoencoders, unsupervised learning, and deep architectures.
\newblock In \emph{JMLR: Workshop and on Unsupervised and Transfer Learning},
  2012.

\bibitem[Bhardwaj et~al.(2017)Bhardwaj, Choudhury, and Scherer]{mp}
Mohak Bhardwaj, Sanjiban Choudhury, and Sebastian Scherer.
\newblock Learning heuristic search via imitation.
\newblock In \emph{CoRL}, 2017.

\bibitem[Bresenham(1965)]{Bresenham}
J.~E. Bresenham.
\newblock Algorithm for computer control of a digital plotter.
\newblock In \emph{IBM System Journal}, 1965.

\bibitem[Cantrell(2000)]{cheby}
Cyrus.~D. Cantrell.
\newblock Modern mathematical methods for physicists and engineers.
\newblock In \emph{Cambridge University Press}, 2000.

\bibitem[Chaplot et~al.(2021)Chaplot, Pathak, and Malik]{transformer_2}
Devendra~Singh Chaplot, Deepak Pathak, and Jitendra Malik.
\newblock Differentiable spatial planning using transformers.
\newblock In \emph{ICML}, 2021.

\bibitem[Chen and McCoy(2019)]{video_games_3}
Zeyuan Chen and Josh McCoy.
\newblock Augmenting character path planning with layered social influences.
\newblock In \emph{AAAI}, 2019.

\bibitem[Choudhury et~al.(2018)Choudhury, Bhardwaj, Arora, Kapoor, Ranade,
  Scherer, and Dey]{sail}
Sanjiban Choudhury, Mohak Bhardwaj, Sankalp Arora, Ashish Kapoor, Gireeja
  Ranade, Sebastian Scherer, and Debadeepta Dey.
\newblock Data-driven planning via imitation learning.
\newblock In \emph{IJRR}, 2018.

\bibitem[Chrestien et~al.(2020)Chrestien, Pevny, Komenda, and
  Edelkamp]{learning_heuristics}
Leah Chrestien, Tomas Pevny, Antonin Komenda, and Stefan Edelkamp.
\newblock A differentiable loss function for learning heuristics in {A}$^\ast$.
\newblock In \emph{NeurIPS}, 2020.

\bibitem[Cohen(2004)]{euclidean}
David Cohen.
\newblock Precalculus: A problems-oriented approach.
\newblock In \emph{Cengage Learning}, 2004.

\bibitem[Courbariaux et~al.(2016)Courbariaux, Hubara, Soudry, El-Yaniv, and
  Bengio]{binary_net}
Matthieu Courbariaux, Itay Hubara, Daniel Soudry, Ran El-Yaniv, and Yoshua
  Bengio.
\newblock Binarized neural networks: Training deep neural networks with weights
  and activations constrained to +1 or -1.
\newblock In \emph{NeurIPS}, 2016.

\bibitem[Cui and Shi(2011)]{video_games_1}
Xiao Cui and Hao Shi.
\newblock Direction oriented pathfinding in video games.
\newblock In \emph{International Journal of Artificial Intelligence and
  Applications}, 2011.

\bibitem[Daniel et~al.(2010)Daniel, Nash, Koenig, and Felner]{theta_star}
Kenny Daniel, Alex Nash, Sven Koenig, and Ariel Felner.
\newblock {Theta}$^\ast$: Any-angle path planning on grids.
\newblock In \emph{JAIR}, 2010.

\bibitem[de~Castro et~al.(2023)de~Castro, Berger, Cantieri, Teixeira, Lima,
  Pereira, and Pinto]{farm_2}
Gabriel G.~R. de Castro, Guido~Szekir Berger, Alvaro Cantieri, Marco
  Antonio~Simões Teixeira, José Lima, Ana~I. Pereira, and Milena~F. Pinto.
\newblock Adaptive path planning for fusing rapidly exploring random trees and
  deep reinforcement learning in an agriculture dynamic environment uavs.
\newblock In \emph{Agriculture}, 2023.

\bibitem[Dechter and Pearl(1985)]{best_first}
Rina Dechter and Judea Pearl.
\newblock Generalized best-first search strategies and the optimality of
  {A$^\ast$}.
\newblock In \emph{JACM}, 1985.

\bibitem[Dijkstra(1959)]{dijkstra}
Edsger~Wybe Dijkstra.
\newblock A note on two problems in connexion with graphs.
\newblock In \emph{Numerische Mathematik}, 1959.

\bibitem[Fan et~al.(2016)Fan, Su, and Guibas]{cdist}
Haoqiang Fan, Hao Su, and Leonidas Guibas.
\newblock A point set generation network for {3D} object reconstruction from a
  single image.
\newblock In \emph{CVPR}, 2016.

\bibitem[Filippis et~al.(2011)Filippis, Guglieri, and Quagliotti]{3d_1}
Luca~De Filippis, Giorgio Guglieri, and Fulvia Quagliotti.
\newblock Path planning strategies for uavs in {3D} environments.
\newblock In \emph{Journal of Intelligent and Robotic Systems}, 2011.

\bibitem[Fishman et~al.(2023)Fishman, Murali, Eppner, Peele, Boots, and
  Fox]{motion_policy}
Adam Fishman, Adithyavairan Murali, Clemens Eppner, Bryan Peele, Byron Boots,
  and Dieter Fox.
\newblock Motion policy networks.
\newblock In \emph{CoRL}, 2023.

\bibitem[Fransen and van Eekelen(2021)]{turning_cost}
Karlijn Fransen and Joost van Eekelen.
\newblock Efficient path planning for automated guided vehicles using
  {A}$^\ast$ ({Astar}) algorithm incorporating turning costs in search
  heuristic.
\newblock In \emph{International Journal of Production Research}, 2021.

\bibitem[Galceran and Carreras(2013)]{ppp_1}
Enric Galceran and Marc Carreras.
\newblock A survey on coverage path planning for robotics.
\newblock In \emph{RAS}, 2013.

\bibitem[Hart et~al.(1968)Hart, Nilsson, and Raphael]{astar}
Peter~E. Hart, Nils~J. Nilsson, and Bertram Raphael.
\newblock A formal basis for the heuristic determination of minimum cost paths.
\newblock In \emph{TSSC}, 1968.

\bibitem[Janson et~al.(2013)Janson, Schmerling, Clark, and
  Pavone]{fast_marching}
Lucas Janson, Edward Schmerling, Ashley Clark, and Marco Pavone.
\newblock Fast marching tree: a fast marching sampling-based method for optimal
  motion planning in many dimensions.
\newblock In \emph{arXiv:1306.3532}, 2013.

\bibitem[Jekel et~al.(2018)Jekel, Venter, Venter, Stander, and Haftka]{area}
Charles~F. Jekel, Gerhard Venter, Martin~P. Venter, Nielen Stander, and
  Raphael~T. Haftka.
\newblock Similarity measures for identifying naterial parameters from
  hysteresis loops using inverse analysis.
\newblock In \emph{International Journal of Material Forming}, 2018.

\bibitem[Karaman et~al.(2011)Karaman, Walter, Perez, Frazzoli, and
  Teller]{anytime_rrt}
Sertac Karaman, Matthew~R. Walter, Alejandro Perez, Emilio Frazzoli, and Seth
  Teller.
\newblock Anytime motion planning using the {RRT$^\ast$}.
\newblock In \emph{ICRA}, 2011.

\bibitem[Kavraki et~al.(1996)Kavraki, Svestka, Latombe, and
  Overmars]{prob_roadmap}
L.~E. Kavraki, P. Svestka, J.~C. Latombe, and M.~H. Overmars.
\newblock Probabilistic roadmaps for path planning in high-dimensional
  configuration spaces.
\newblock In \emph{TRA}, 1996.

\bibitem[Kirilenko et~al.(2023)Kirilenko, Andreychuk, Panov, and
  Yakovlev]{trans_path}
Daniil Kirilenko, Anton Andreychuk, Aleksandr Panov, and Konstantin Yakovlev.
\newblock {TransPath}: Learning heuristics for grid-based pathfinding via
  transformers.
\newblock In \emph{AAAI}, 2023.

\bibitem[Kolmogorov(2006)]{trws}
Vladimir Kolmogorov.
\newblock Convergent tree-reweighted message passing for energy minimization.
\newblock In \emph{Transactions on Pattern Analysis and Machine Intelligence},
  2006.

\bibitem[Krizhevsky et~al.(2012)Krizhevsky, Sutskever, and Hinton]{cnn}
Alex Krizhevsky, Ilya Sutskever, and Geoffrey~E. Hinton.
\newblock {ImageNet} classification with deep convolutional neural networks.
\newblock In \emph{NeurIPS}, 2012.

\bibitem[Kuffner and LaValle(2000)]{rrt_connect}
J.J. Kuffner and S.M. LaValle.
\newblock {RRT}-connect: An efficient approach to single-query path planning.
\newblock In \emph{ICRA}, 2000.

\bibitem[LaValle(1998)]{rrt}
Steven~M. LaValle.
\newblock Rapidly-exploring random trees: A new tool for path planning.
\newblock In \emph{Technical Report}, 1998.

\bibitem[Lee et~al.(2018)Lee, Parisotto, Chaplot, Xing, and
  Salakhutdinov]{gt_path_dijkstra}
Lisa Lee, Emilio Parisotto, Devendra~Singh Chaplot, Eric Xing, and Ruslan
  Salakhutdinov.
\newblock Gated path planning networks.
\newblock In \emph{ICML}, 2018.

\bibitem[Li et~al.(2023{\natexlab{a}})Li, Qin, Han, and Lu]{gradsmqga}
Hui Li, Huiping Qin, Zi'ao Han, and Kai Lu.
\newblock Vehicle path planning based on gradient statistical mutation quantum
  genetic algorithm.
\newblock In \emph{International Journal of Advanced Computer Science and
  Applications}, 2023{\natexlab{a}}.

\bibitem[Li et~al.(2023{\natexlab{b}})Li, Zhang, Hu, Fu, Liao, and Yu]{RJAStar}
Jian Li, Weijian Zhang, Yating Hu, Shengliang Fu, Changyi Liao, and Weilin Yu.
\newblock {RJA-Star} algorithm for {UAV} path planning based on improved
  {R5DOS} model.
\newblock In \emph{Applied Sciences}, 2023{\natexlab{b}}.

\bibitem[Moore(1959)]{bfs}
Edward~F. Moore.
\newblock The shortest path through a maze.
\newblock In \emph{International Symposium on the Theory of Switching}, 1959.

\bibitem[Overmars(2005)]{video_games_2}
M. Overmars.
\newblock Path planning for games.
\newblock In \emph{Computer Science}, 2005.

\bibitem[Pak et~al.(2022)Pak, Kim, Park, and Son]{farm_1}
Jeonghyeon Pak, Jeongeun Kim, Yonghyun Park, and Hyoung~Il Son.
\newblock Field evaluation of path-planning algorithms for autonomous mobile
  robot in smart farms.
\newblock In \emph{IEEE Access}, 2022.

\bibitem[Pearl and Kim(1982)]{focal_search}
Judea Pearl and Jin~H. Kim.
\newblock Studies in semi-admissible heuristics.
\newblock In \emph{TPAMI}, 1982.

\bibitem[Peyre and Cohen(2006)]{3d_app}
G. Peyre and L.~D. Cohen.
\newblock Landmark-based geodesic computation for heuristically driven path
  planning.
\newblock In \emph{CVPR}, 2006.

\bibitem[Rahim et~al.(2018)Rahim, Abdullah, Nurarif, Ramadhan, Anwar, Dahria,
  Nasution, Diansyah, and Khairani]{breadth_first}
Robbi Rahim, Dahlan Abdullah, Saiful Nurarif, Mukhlis Ramadhan, Badrul Anwar,
  Muhammad Dahria, Surya~Darma Nasution, Tengku~Mohd Diansyah, and Mufida
  Khairani.
\newblock Breadth first search approach for shortest path solution in cartesian
  area.
\newblock In \emph{JPCS}, 2018.

\bibitem[Robicquet et~al.(2016)Robicquet, Sadeghian, Alahi, and Savarese]{sdd}
Alexandre Robicquet, Amir Sadeghian, Alexandre Alahi, and Silvio Savarese.
\newblock Learning social etiquette: Human trajectory understanding in crowded
  scenes.
\newblock In \emph{ECCV}, 2016.

\bibitem[Shea and Nash(2015)]{cnn_2}
Keiron~O’ Shea and Ryan Nash.
\newblock An introduction to convolutional neural networks.
\newblock In \emph{arXiv:1511.08458}, 2015.

\bibitem[Sturtevant(2012)]{city_street}
Nathan Sturtevant.
\newblock Benchmarks for grid-based pathfinding.
\newblock In \emph{Transactions on Computational Intelligence and AI in Games},
  2012.

\bibitem[Takahashi et~al.(2019)Takahashi, Sun, Tian, and Wang]{deep_heuristic}
Takeshi Takahashi, He Sun, Dong Tian, and Yebin Wang.
\newblock Learning heuristic functions for mobile robot path planning using
  deep neural networks.
\newblock In \emph{ICAPS}, 2019.

\bibitem[Tarjan(1971)]{depth_first}
Robert Tarjan.
\newblock Depth-first search and linear graph algorithms.
\newblock In \emph{Annual Symposium on Switching and Automata Theory}, 1971.

\bibitem[Thoma et~al.(2019)Thoma, Paudel, Chhatkuli, Probst, and Gool]{ppp_2}
Janine Thoma, Danda~Pani Paudel, Ajad Chhatkuli, Thomas Probst, and Luc~Van
  Gool.
\newblock Mapping, localization and path planning for image-based navigation
  using visual features and map.
\newblock In \emph{CVPR}, 2019.

\bibitem[Tian et~al.(2023)Tian, Mo, Ma, Xiao, Jia, Lan, and
  Zhang]{UAV_multi_obj}
Haoxin Tian, Zhenjie Mo, Chenyang Ma, Junqi Xiao, Ruichang Jia, Yubin Lan, and
  Yali Zhang.
\newblock Design and validation of a multi-objective waypoint planning
  algorithm for {UAV} spraying in orchards based on improved ant colony
  algorithm.
\newblock In \emph{Frontiers in Plant Science}, 2023.

\bibitem[Vaswani et~al.(2017)Vaswani, Shazeer, Parmar, Uszkoreit, Jones, Gomez,
  Kaiser, and Polosukhin]{transformer}
Ashish Vaswani, Noam Shazeer, Niki Parmar, Jakob Uszkoreit, Llion Jones,
  Aidan~N. Gomez, Lukasz Kaiser, and Illia Polosukhin.
\newblock Attention is all you need.
\newblock In \emph{NeurIPS}, 2017.

\bibitem[Vlastelica et~al.(2020)Vlastelica, Paulus, Musil, Martius, and
  Rolinek]{bbox}
Marin Vlastelica, Anselm Paulus, Vít Musil, Georg Martius, and Michal Rolinek.
\newblock Differentiation of blackbox combinatorial solvers.
\newblock In \emph{ICLR}, 2020.

\bibitem[Wang et~al.(2022)Wang, Lou, Jing, Wang, Liu, and Liu]{EBSAStar}
Huanwei Wang, Shangjie Lou, Jing Jing, Yisen Wang, Wei Liu, and Tieming Liu.
\newblock The {EBS-A}$^\ast$ algorithm: An improved {A}$^\ast$ algorithm for
  path planning.
\newblock In \emph{PLoS ONE}, 2022.

\bibitem[Wang et~al.(2020)Wang, Luo, Han, Chen, Liang, and Zheng]{random_tree}
Xinda Wang, Xiao Luo, Baoling Han, Yuhan Chen, Guanhao Liang, and Kailin Zheng.
\newblock Collision-free path planning method for robots based on an improved
  rapidly-exploring random tree algorithm.
\newblock In \emph{Applied Sciences}, 2020.

\bibitem[Xu et~al.(2020)Xu, Ajanthan, and Hartley]{mplayers}
Zhiwei Xu, Thalaiyasingam Ajanthan, and Richard Hartley.
\newblock Fast and differentiable message passing on pairwise markov random
  fields.
\newblock In \emph{Asian Conference on Computer Vision}, 2020.

\bibitem[Yakovlev et~al.(2015)Yakovlev, Baskin, and
  Hramoin]{grid_angle_constrained}
Konstantin Yakovlev, Egor Baskin, and Ivan Hramoin.
\newblock Grid-based angle-constrained path planning.
\newblock In \emph{AAAI}, 2015.

\bibitem[Yan et~al.(2014)Yan, Liu, and Xiao]{3d_2}
Fei Yan, Yisha Liu, and Jizhong Xiao.
\newblock Path planning in complex {3D} environments using a probabilistic
  roadmap method.
\newblock In \emph{IJAC}, 2014.

\bibitem[Yonetani et~al.(2021)Yonetani, Taniai, Barekatain, Nishimura, and
  Kanezaki]{neural_a_star}
Ryo Yonetani, Tatsunori Taniai, Mohammadamin Barekatain, Mai Nishimura, and
  Asako Kanezaki.
\newblock Path planning using neural {A}$^\ast$ search.
\newblock In \emph{ICML}, 2021.

\bibitem[Zang et~al.(2023)Zang, Yin, Xiao, Zonouz, and Yuan]{graphcnn}
Xiao Zang, Miao Yin, Jinqi Xiao, Saman Zonouz, and Bo Yuan.
\newblock Graphmp: Graph neural network-based motion planning with efficient
  graph search.
\newblock In \emph{NeurIPS}, 2023.

\bibitem[Zhang et~al.(2021)Zhang, Wu, Shen, and Li]{auto_land_vehicle}
Jing Zhang, Jun Wu, Xiao Shen, and Yunsong Li.
\newblock Autonomous land vehicle path planning algorithm based on improved
  heuristic function of {A-Star}.
\newblock In \emph{IJARS}, 2021.

\bibitem[Zhen et~al.(2023)Zhen, Gu, Shi, and Suo]{water_traffic}
Rong Zhen, Qiyong Gu, Ziqiang Shi, and Yongfeng Suo.
\newblock An improved {A-Star} ship path-planning algorithm considering
  current, water depth, and traffic separation rules.
\newblock In \emph{JMSE}, 2023.

\end{thebibliography}
\bibliographystyle{ieeenat_fullname}
}

\clearpage
\setcounter{table}{5}
\setcounter{figure}{2}
\setcounter{algorithm}{1}

\appendix
\section*{Appendix}
\section{Learned Weights}
\label{sec:weights}

To highlight the importance and validation of the path angular freedom (PAF),
we show the learned $\alpha$ in the main paper.
Here, we provide all the learned weights in Table~\ref{tb:learned_weights}.
\begin{table}[!ht]
\centering
\setlength{\tabcolsep}{7pt}
\resizebox{0.47\textwidth}{!}{
\begin{tabular}{ll|rrr}
\Xhline{2\arrayrulewidth}
\multicolumn{1}{c}{\textbf{Dataset}}
& \multicolumn{1}{c|}{\textbf{Method}}
& \multicolumn{1}{c}{$\boldsymbol{\alpha}$}
& \multicolumn{1}{c}{$\boldsymbol{\lambda}$}
& \multicolumn{1}{c}{$\boldsymbol{\kappa}$} \\
\Xhline{2\arrayrulewidth}
\multirow{3}{*}{MPD} & DA\astar-min & 100.0 & 19.9 & 29.6 \\
& DA\astar-max & 0.0 & 52.5 & 36.2 \\
& DA\astar-mix & 33.4 & 66.0 & 75.3 \\
\hdashline
\multirow{3}{*}{TMPD} & DA\astar-min & 100.0 & 17.2 & 65.9 \\
& DA\astar-max & 0.0 & 32.1 & 0.1 \\
& DA\astar-mix & 76.9 & 48.1 & 83.8 \\
\hdashline
\multirow{3}{*}{CSM} & DA\astar-min & 100.0 & 47.3 & 100.0 \\
& DA\astar-max & 0.0 & 32.1 & 0.1 \\
& DA\astar-mix & 68.7 & 51.7 & 76.8 \\
\hline
\multirow{4}{*}{Aug-TMPD} & DA\astar & 96.4 & 63.3 & 97.8 \\
& DA\astar-path & 58.7 & 55.2 & 56.3 \\
& DA\astar-mask & 69.6 & 50.5 & 77.4 \\
& DA\astar-weight & 62.2 & 49.9 & 70.3 \\
\hline
\multirow{3}{*}{Warcraft} & DA\astar-min & 100.0 & 75.4 & 16.7 \\
& DA\astar-max & 0.0 & 78.8 & 43.9 \\
& DA\astar-mix & 27.7 & 80.5 & 74.4 \\
\hdashline
\multirow{3}{*}{Pok\'emon} & DA\astar-min & 100.0 & 78.5 & 22.8 \\
& DA\astar-max & 0.0 & 83.4 & 44.0 \\
& DA\astar-mix & 32.3 & 83.2 & 70.2 \\
\hline
\multirow{3}{*}{SDD-intra} & DA\astar-min & 100.0 & 53.5 & 0.2 \\
& DA\astar-max & 0.0 & 73.4 & 68.4 \\
& DA\astar-mix & 9.5 & 77.9 & 91.4 \\
\hdashline
\multirow{3}{*}{SDD-inter} & DA\astar-min & 100.0 & 29.6 & 3.0 \\
& DA\astar-max & 0.0 & 69.0 & 57.7 \\
& DA\astar-mix & 17.4 & 73.0 & 84.1 \\
\Xhline{2\arrayrulewidth}
\end{tabular}
}
\caption{Learned weights, scaled by $\times$100 for high readability.
We set $\alpha=1$ for DA\astar-min and $\alpha=0$ for DA\astar-max.
We report the mean values where multiple training seeds are applied.
}
\label{tb:learned_weights}
\end{table}

\section{Relation of $\boldsymbol{\alpha}$ and Predicted Paths}
\label{sec:relation}

In Eq.~\eqref{eq:PAF}, a large $\alpha$ aims to minimize the path angle to often achieve linear paths, while a small one maximizes it for smooth paths.
The learned $\alpha=0.69$ in CSM encourages small path angles for short paths, such as linear path segments. For example, in Fig.~\ref{fig:examples}(\iccvcolor{a}), the path consists essentially of 3 main linear segments by DA\astar. This inspection also explains Fig.~\ref{fig:examples}(\iccvcolor{b}) for the learned $\alpha=0.62$ by DA\astar.

Meanwhile, unlike these binary maps in Figs.~\ref{fig:examples}(\iccvcolor{a})--(\iccvcolor{b}) where obstacles and accessible areas are provided in the datasets, video-game maps and SDD have more complex scenarios and have to learn cost maps in Figs.~\ref{fig:examples}(\iccvcolor{c})--(\iccvcolor{f}).
These learned costs are often continuous on the edges of objects such as rocks, lakes, and roundabouts.
Thus, smooth paths can generally be achieved by increasing the path angle. This aligns with the small $\alpha$, 0.28 and 0.32 on video-game datasets and particularly 0.1 and 0.17 on SDD where realistic road scenes require smaller $\alpha$ to obtain smoother paths considering driving safety.

\section{Training Procedure}
\label{sec:tranining}
In addition to the pathfinding of DA\astar~in Alg.~\ref{alg:daa_code}, we also detail the training procedure with path loss and the combination of path loss and PPM loss in Alg.~\ref{alg:daa_train}.

\begin{algorithm}[t]
\caption{Training Procedure of Deep Angular \astar}
\label{alg:daa_train}
\textbf{Input}: A dataset $\mathcal D=\{\mathcal{I}_i, s_i, t_i, \hat{\mathcal{P}}_i, \hat{\boldsymbol{\theta}}_i\}$ with a map image $\mathcal{I}_i$ containing $N$ pixels, a source node $s_i$, a target node $t_i$, a reference path $\hat{\mathcal{P}}_i$, and a reference PPM $\hat{\boldsymbol{\theta}}_i$ (only required by PPM loss) for all $i \in \{1, ..., M\}$ given $M$ samples, a mask threshold $\mathcal{T}=0.5$, learning rate $l$, and loss weights $w_1$ and $w_2$ for path loss and PPM loss, respectively.\\
\textbf{Parameter}: A cost-map encoder network $f(\mc{I},s,t)$, containing weights $\mathbf{w}$ and bias $\mathbf{b}$, and term weights $\{\alpha, \lambda, \kappa\}$.\\
\textbf{Note}: On \textit{Aug-TMPD}, we set $w_1=w_2=1$ for DA\astar, $w_1=1$ and $w_2=0$ for DA\astar-path, $w_1=w_2=1$ for DA\astar-mask with a cost-map mask, and $w_1=10$ and $w_2=1$ for DA\astar-weight. \textit{On the other datasets}, we set $w_1=1$ and $w_2=0$ since only the path loss is required.

\textbf{Procedure}:
\begin{algorithmic}[1]
\STATE Initialize $\alpha=0.5$, $\lambda=0.5$, and $\kappa=1$.
\FOR{$i \in \{1, ..., M\}$}
    \STATE Compute a cost map $\boldsymbol{\theta}_i=f(\mathcal{I}_i, s_i, t_i)$.
    \STATE Compute a confidence list $\boldsymbol{p}=\{p_k\}$ in Eq.~(\iccvcolor{9}) for all $k \in \mathcal{M}_c$ by Step 1, Alg.~\iccvcolor{1} given inputs $\{\boldsymbol{\theta}_i, s_i, t_i\}$.
    \STATE Form a reference path map $\hat{\mathcal{Y}}_i \in \mathcal{B}^N$ with 1 for all nodes in $\hat{\mathcal{P}}_i$ and 0 otherwise.
    \STATE Form a search history map $\mathcal{Y} \in \mathcal{B}^N$ with $(1-p_k)_{\text{detach}} +p_k$ for all $k \in \mathcal{M}_c$ using discretized activation\footnotemark~and 0 otherwise.
    \STATE Compute path loss $\mathcal{L}^p_i=\vert| \mathcal{Y} - \hat{\mathcal{Y}}_i \vert|_1$.
    \STATE For DA\astar-mask, compute a cost-map mask $\boldsymbol{m}=\mathbb{1}[\boldsymbol{\theta}_i \geq \mathcal{T}]$, then update $\boldsymbol{\theta}_i \leftarrow \boldsymbol{\theta}_i \odot \boldsymbol{m}$ and $\hat{\boldsymbol{\theta}}_i \leftarrow \hat{\boldsymbol{\theta}}_i \odot \boldsymbol{m}$ using Hadamard product.
    \STATE Compute PPM loss $\mathcal{L}^c_i=\vert| \boldsymbol{\theta}_i - (1-\hat{\boldsymbol{\theta}}_i) \vert|_2$.
    \STATE Compute a weighted loss $\mathcal{L}_i = w_1 \mathcal{L}^p_i + w_2 \mathcal{L}^c_i$.
\ENDFOR
\STATE Compute the average loss $\mathcal{L}=\frac{1}{M} \sum^M_{i=1} \mathcal{L}_i$.
\STATE Compute the gradients of learnable parameters, $\nabla_{\boldsymbol{w}} \mathcal{L}$, $\nabla_{\boldsymbol{b}} \mathcal{L}$, $\nabla_{\alpha} \mathcal{L}$, $\nabla_{\lambda} \mathcal{L}$, and $\nabla_{\kappa} \mathcal{L}$, by backpropagting $\mathcal{L}$.
\STATE Update parameters $\boldsymbol{w} \leftarrow \boldsymbol{w} - l \nabla_{\boldsymbol{w}} \mathcal{L}$, $\boldsymbol{b} \leftarrow \boldsymbol{b} - l \nabla_{\boldsymbol{b}} \mathcal{L}$, $\alpha \leftarrow \alpha - l \nabla_{\alpha} \mathcal{L}$, $\lambda \leftarrow \lambda - l \nabla_{\lambda} \mathcal{L}$, and $\kappa \leftarrow \kappa - l \nabla_{\kappa} \mathcal{L}$.
\STATE Repeat Lines 2-14 till the training loss converges.
\end{algorithmic}
\end{algorithm}
\footnotetext{The discretized activation disables gradient accumulation by $(\cdot)_{\text{detach}}$, see \href{https://github.com/omron-sinicx/neural-astar/blob/minimal/src/neural_astar/planner/differentiable_astar.py\#74}{Line 74} of \astar. This is equivalent to computing the gradients from $p_k$.}

\section{Ethical and Societal Impact}
\label{sec:impact}

This work aims to automate effective vision-based path planning through end-to-end learning of path shortening and smoothing.
While our DA\astar~promises significant improvements in imitating human and machine demonstrations, it also raises potential ethical and societal concerns such as auto-driving safety in complex environments, model robustness under attacks, and energy consumption in model training.
For instance, under certain attacks in a system such as ADAS, 
$\alpha$ can be maliciously changed to 1, where predicted paths are forced to be linear crossing inaccessible areas say roundabouts or unsafe zigzags causing traffic chaos.

However, according to the trade-off between searching efficiency and path optimality, the search cost 0.5\%--6.4\% is minor.
We also reveal the necessity of only binary path labelling for supervised learning with more accessible path reference rather than PPM reference.
We aim to further mitigate these issues by developing fair, transferable, robust, and more efficient algorithms in future work.
Our commitment is to ensure that autonomous path planning is socially responsible and environmentally sustainable.

\end{document}